\documentclass{article}
\usepackage[dvips]{graphicx}
\usepackage{amsmath,amssymb}
\usepackage{fancybox, enumerate}
\usepackage{bm}
\usepackage{my-style}
\allowdisplaybreaks[4]
\newcommand{\myhrulefill}{\leavevmode%
\leaders\hrule depth-2.1pt height 2.5pt\hfill\kern0pt
}
\newcommand{\Mychrulefill}{\leavevmode%
\leaders\hrule depth-2.1pt height 3.5pt\hfill\kern0pt
}

\graphicspath{{fig/}}

\begin{document}
\begin{center}
\Large{\textsc{Consistent Batch Normalization for Weighted Loss in Imbalanced-Data Environment}}\\
\end{center}
\begin{center}
Muneki Yasuda and Yeo Xian En
\\
Graduate School of Science and Engineering, Yamagata University, Japan\\
\vspace{2mm}
Seishirou Ueno\\
FUJITSU BROAD SOLUTION \& CONSULTING Inc., Japan
\end{center}

\begin{abstract}
In this study, classification problems based on feedforward neural networks in a data-imbalanced environment are considered. 
Learning from an imbalanced dataset is one of the most important practical problems in the field of machine learning. 
A weighted loss function (WLF) based on a cost-sensitive approach is a well-known and effective method for imbalanced datasets. 
A combination of WLF and batch normalization (BN) is considered in this study. 
BN is considered as a powerful standard technique in the recent developments in deep learning. 
A simple combination of both methods leads to a size-inconsistency problem due to a mismatch between the interpretations of the effective size of the dataset in both methods. 
A simple modification to BN, called weighted BN (WBN), is proposed to correct the size mismatch. 
The idea of WBN is simple and natural. 
The proposed method in a data-imbalanced environment is validated using numerical experiments.
\end{abstract}

\section{Introduction}

Learning from an imbalanced dataset is one of the most important and practical problems in the field of machine learning, 
and it has been actively studied~\cite{ReviewIBD2017,He2009}. 
This study focuses on classification problems based on feedforward neural networks with imbalanced datasets. 
Several methods were proposed for these problems~\cite{He2009}, for example, under/over-sampling, synthetic minority oversampling technique (SMOTE)~\cite{SMOTE2002}, 
and cost-sensitive (CS) method. A CS-based method is considered in this study.
A simple type of CS method was developed by assigning weights to each point of data loss in a loss function~\cite{He2009,CSM2010}. 
The weights in the CS method can be viewed as importances of corresponding data points. 
As discussed in Section \ref{sec:WLF}, the weighting changes the interpretation of the effective size of each data point in the resultant \textit{weighted loss function} (WLF). 
In an imbalanced dataset, the weights of data points that belong to the majority classes are often set to be smaller than those that belong to the minority classes. 
For imbalanced datasets, two different effective settings of the weights are known: \textit{inverse class frequency} (ICF)~\cite{CSM2016,CSM2017,CSM2018} 
and \textit{class-balance loss} (CBL)~\cite{CBL2019}.

\textit{Batch normalization} (BN) is a powerful regularization method for neural networks~\cite{BN2015}; 
moreover is one of the most important techniques in deep learning~\cite{DL2016}. 
In BN, the affine signals from the lower layer are normalized over a specific mini-batch. 
A simple combination of WLF and BN causes a \textit{size-inconsistency problem}, 
which degrades the classification performance. 
As mentioned above, the interpretation of the effective size of data points is changed in WLF. 
However, in the standard scenario of BN, one data point is counted as just one data point in calculations. 
This inconsistency results in the size-inconsistency problem.   

The aim of this study is not to identify a better learning method in a data-imbalanced environment, 
but to resolve the size-inconsistency problem in the simple combination of WLF and BN.  
In this study, a consistent BN is proposed to resolve the size-inconsistency problem.   
This paper is an extension of our previous study~\cite{NOLTA2019} with certain new improvements: 
(i) the theory is improved (i.e., the definition of signal variance in the proposed method is improved) 
and (ii) more experiments were conducted (i.e., for new datasets and for a new WLF setting).

The remainder of this paper is organized as follows. 
In Section \ref{sec:WLF}, WLF and the two different settings of the weights, i.e., ICF and CBL, for a data-imbalanced environment are briefly explained. 
In Section \ref{sec:BN}, a brief explanation of BN is presented. 
In this section, the performances of a simple combination of WLF and BN in certain data-imbalanced environments are shown 
using numerical experiments conducted with MNIST, Fashion-MNIST~\cite{fashion-MNIST2017}, and CIFAR-10 datasets.  
The simple combination method marginally improves the classification performance in the data-imbalanced environments; however, this improvement is insufficient. 
This is caused by the size-inconsistency problem mentioned above.
In Section \ref{sec:WBN}, a consistent BN, which can resolve the size-inconsistency problem, is proposed  
and the proposed method is validated using numerical experiments.
As shown in the numerical experiments in Section \ref{sec:experiment2}, the proposed method significantly improves the classification performance.
Finally, the summary and certain future works are presented in Section ~\ref{sec:summary}.
In the following sections, the definition sign ``$:=$'' is used only when the expression does not change throughout the paper. 
In the definitions of parameters that are redefined in other sections of the paper, the definition sign is not used.   

\section{Weighted Loss Function in Imbalanced Data Environment}
\label{sec:WLF}

Consider a classification model that classifies an $n$-dimensional input vector $\bm{x} := (x_1, x_2,\ldots, x_n)^{\mrm{T}}$ into $K$ different classes 
$C_1, C_2, \ldots, C_K$. 
It is convenient to use a 1-of-$K$ vector (or one-hot vector) to identify each class~\cite{Bishop2006}. 
Here, each class corresponds to the $K$ dimensional vector $\bm{t} := (t_1, t_2,\ldots, t_K)^{\mrm{T}}$ having elements $t_k \in \{0,1\}$ 
and $\sum_{k = 1}^K t_k = 1$, i.e., a vector in which only one element is one and the remaining elements are zero. 
When $t_k = 1$, $\bm{t}$ indicates class $C_k$. 
For the sake of simplicity, the 1-of-$K$ vector whose $k$th element of which is one is denoted by $\bm{1}_k$; thus, $\bm{t} \in T_K := \{\bm{1}_k \mid k = 1,2,\ldots,K\}$. 
Thus, $\bm{t} = \bm{1}_k$ corresponds to class $C_k$. 
The size of the data points belonging to class $C_k$ is denoted by $N_k$, defined as $N_k := \sum_{\mu \in \Omega}\delta(\bm{1}_k, \mbf{t}_{\mu})$, 
where $\delta(\bm{a},\bm{b})$ is the Kronecker delta. 
From the definition, $\sum_{k=1}^K N_k = N$ is obtained.

Let us consider a training dataset comprising of $N$ data points, i.e.,  
$\mcal{D}:= \{\{\mbf{x}_{\mu}, \mbf{t}_{\mu}\} \mid \mu \in \Omega :=\{ 1,2,\ldots,N\}\}$, 
where $\mbf{x}_{\mu} := (\mrm{x}_{1,\mu}, \mrm{x}_{1,\mu},\ldots, \mrm{x}_{n,\mu})^{\mrm{T}}$ 
and $\mbf{t}_{\mu}:= (\mrm{t}_{1,\mu}, \mrm{t}_{2,\mu},\ldots, \mrm{t}_{K,\mu})^{\mrm{T}} \in T_K$ are the $\mu$th input 
and the corresponding target-class label represented by the 1-of-$K$ vector, respectively.   
In the standard machine-learning scenario, a loss function given by   
\begin{align}
\mcal{L}(\theta):= \frac{1}{N}\sum_{\mu \in \Omega} f(\mbf{x}_{\mu}, \mbf{t}_{\mu}; \theta)
\label{eqn:LossFunction}
\end{align}
is minimized with respect to $\theta$ using an appropriate back-propagation algorithm, 
where $f(\mbf{x}_{\mu}, \mbf{t}_{\mu}; \theta)$ denotes the data loss (e.g., the cross-entropy loss) for the $\mu$th data point $\{\mbf{x}_{\mu}, \mbf{t}_{\mu}\}$ 
and $\theta$ denotes the set of learning parameters of the classification model. 

WLF is introduced as 
\begin{align}
\mcal{L}_{\mrm{w}}(\theta):= \frac{1}{Z}\sum_{ \mu \in \Omega} w_{\mu} f(\mbf{x}_{\mu}, \mbf{t}_{\mu}; \theta),
\label{eqn:WeightedLossFunction}
\end{align}
where $w_{\mu} >  0$ is the weight of the $\mu$th data point and $Z := \sum_{\mu \in \Omega}w_{\mu}$ is the normalization factor. 
It can be viewed as a simple type of CS approach~\cite{He2009,CSM2010}. 
In WLF, the $\mu$th data point $\{\mbf{x}_{\mu}, \mbf{t}_{\mu}\}$ is replicated to $w_{\mu}$ data points. 
Hence, the relative occupancy (or the effective size) of the data points in each class is changed according to the weights. 
In WLF, the ``effective'' size of $N_k$ is then expressed by 
\begin{align}
N_k^{\mrm{eff}} := \sum_{\mu \in \Omega} w_{\mu}\delta(\bm{1}_k, \mbf{t}_{\mu}).
\label{eqn:Effective_DataSize}
\end{align}
In general, 
\begin{align}
Z = \sum_{k = 1}^K \sum_{\mu \in \Omega_k}w_{\mu} = \sum_{k = 1}^KN_k^{\mrm{eff}}.
\label{eqn:Z&Neff}
\end{align}
The effect of WLF presented here is essentially the same as that of a simple under- or over-sampling method. 
The over-sampling method causes an increase in the size of dataset, while WLF dose not cause this. 
This is the major merit of WLF when compared to the simple over-sampling method.

The weights $\bm{w}= (w_1, w_2,\ldots, w_{\Omega})^{\mrm{T}}$ should be individually set according to the tasks. 
For an imbalanced dataset (i.e., certain values of $N_k$ are very large (\textit{majority} classes) and certain values are very small (\textit{minority} classes)), 
there are two known settings: ICF~\cite{CSM2016,CSM2017,CSM2018} and CBL~\cite{CBL2019}.  
In the following sections, the two settings are briefly explained.

\subsection{Inverse class frequency}
\label{sec:ICF}

\begin{figure}[tb]
\centering
\includegraphics[height=1.5cm]{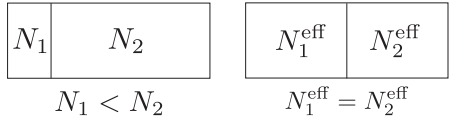}
\caption{Example of a two-class case: $N_1 = 1$ (\textit{minority} class) and $N_2 = 99$ (\textit{majority} class). 
The effective sizes of $N_k^{\mrm{eff}}$ are the same: $N_1^{\mrm{eff}} = N_2^{\mrm{eff}} = N = 100$ in ICF.}
\label{fig:data_balance}
\end{figure}

Here, let us assume that $\mcal{D}$ is an imbalanced dataset, which implies that the sizes of $N_k$s are imbalanced. 
In ICF, the weights are set to
\begin{align}
w_{\mu} = w_{\mu}^{\mrm{icf}}:=\Big(\frac{1}{N}\sum_{k=1}^K N_k \delta(\bm{1}_k,\mbf{t}_{\mu})\Big)^{-1},
\label{eqn:InverseWeight}
\end{align}
where $N_k \geq 1$ is assumed.
The ICF weights $\bm{w}^{\mrm{icf}}$ effectively correct the imbalance in the loss function due to the imbalanced dataset. 
In WLF with ICF, the $\mu$th data point that belongs to class $\mcal{C}_k$ is replicated to $w_{\mu}^{\mrm{icf}} = N / N_{k}$ data points.
Therefore, in this setting, the effective sizes of $N_k$ are equal to $N$ for any $k$: 
\begin{align*}
N_k^{\mrm{eff}} = N \sum_{\mu \in \Omega} \frac{\delta(\bm{1}_k, \mbf{t}_{\mu})}{\sum_{l=1}^K N_l \delta(\bm{1}_l,\mbf{t}_{\mu})}
=N \sum_{\mu \in \Omega_k} \frac{1}{\sum_{l=1}^K N_l \delta(\bm{1}_l,\bm{1}_k)} = N,
\end{align*}
where $\Omega_k :=\{ \mu \mid \mbf{t}_{\mu} = \bm{1}_k ,\,\mu \in \Omega\} \subseteq \Omega$ and $|\Omega_k| = N_k$.
This implies that the effective sizes of $N_k$ (i.e., $N_k^{\mrm{eff}}$) are completely balanced in WLF (see Figure \ref{fig:data_balance}).
From equation (\ref{eqn:Z&Neff}), $Z = KN$ in ICF.

\subsection{Class-balanced loss}
\label{sec:CBL}

Recently, a new type of WLF (referred to as CBL), which is effective for imbalanced datasets, was proposed~\cite{CBL2019}.   
In CBL, the weights in equation (\ref{eqn:WeightedLossFunction}) are set to 
\begin{align}
w_{\mu}=w_{\mu}^{\mrm{cbl}}(\beta):=\frac{1-\beta}{1-\beta^{\rho_{\mu}}},
\label{eqn:CBL}
\end{align}
for $\beta \in [0,1)$, where $\rho_{\mu}:=\sum_{k=1}^K N_k \delta(\bm{1}_k,\mbf{t}_{\mu})$. 
In CBL, $N_k \geq 1$ is also assumed.
The CBL weights $\bm{w}^{\mrm{cbl}}(\beta)$ correct the effective sizes of data points by considering the overlaps of data points.
Here, $\beta$ is a hyperparameter that smoothly connects the loss function in equation (\ref{eqn:LossFunction}) 
and WLF with ICF. 
WLF with CBL is equivalent to equation (\ref{eqn:LossFunction}) when $\beta = 0$, 
and it is equivalent to WLF with ICF when $\beta \to 1$, except for the difference in the constant factor:   
\begin{align*}
\lim_{\beta \to 1}w_{\mu}^{\mrm{cbl}}(\beta) = \rho_{\mu}^{-1} = \frac{w_{\mu}^{\mrm{icf}}}{N}.
\end{align*} 
Hence, CBL can be viewed as an extension of ICF.
In Reference \cite{CBL2019}, it was recommended that $\beta$ should have a value close to one (e.g., approximately $\beta = (N -1)/ N$). 
In CBL, the effective sizes of $N_k$ are given by  
\begin{align}
N_k^{\mrm{eff}}=\sum_{\mu \in \Omega} w_{\mu}^{\mrm{cbl}}\delta(\bm{1}_k, \mbf{t}_{\mu})  = \frac{N_k(1-\beta)}{1-\beta^{N_k}}.
\label{eqn:Nk-eff_CBL}
\end{align}
From equations (\ref{eqn:Z&Neff}) and (\ref{eqn:Nk-eff_CBL}), 
\begin{align*}
Z = (1-\beta)\sum_{k=1}^K \frac{N_k}{1-\beta^{N_k}} 
\end{align*}
in CBL.

As mentioned, $\bm{w}^{\mrm{cbl}}(\beta) \propto \bm{w}^{\mrm{icf}}$ in the limit of $\beta \to 1$; 
therefore, the values of $N_k^{\mrm{eff}}$ are completely balanced in this limit.
In fact, $N_k^{\mrm{eff}} = 1$ for all $k$ when $\beta \to 1$. 
The complete balance gradually begins to break as $\beta$ decreases. 
When $\Delta := 1 - \beta > 0$ is very small (i.e., $\beta$ is very close to one), equation (\ref{eqn:Nk-eff_CBL}) is expanded as
\begin{align*}
N_k^{\mrm{eff}}=1 + \frac{N_k - 1}{2}\Delta + \frac{N_k^2 - 1}{12}\Delta^2 +  O(\Delta^3).
\end{align*}
Therefore, when $N_l \neq N_k$, 
\begin{align*}
\frac{N_l^{\mrm{eff}} - N_k^{\mrm{eff}}}{N_l - N_k} \approx O(\Delta).
\end{align*}
This implies that $N_k^{\mrm{eff}}$ and $N_l^{\mrm{eff}}$ are almost balanced when compared to $N_k$ and $N_l$, when $\Delta$ is very small.

\section{Batch Normalization}
\label{sec:BN}

\begin{figure}[tb]
\centering
\includegraphics[height=4cm]{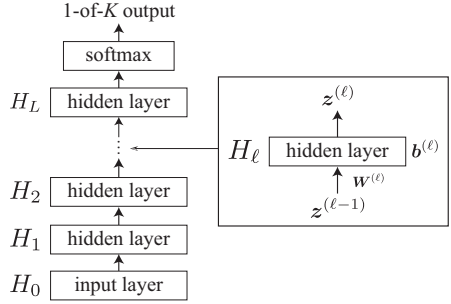}
\caption{Schematic illustration of the feedforward propagation for classification.}
\label{fig:DNN}
\end{figure}

Consider a standard feedforward neural network for classification whose $\ell$th layer consists of $H_{\ell}$ units ($\ell = 0,1,\ldots, L$). 
The zeroth layer ($\ell = 0$) is the input layer, i.e., $H_0 = n$, and the network output, $\bm{t} \in T_K$, is determined from the output signals of the $L$th layer.  
In the standard scenario for the feedforward propagation of input $\mbf{x}^{(\mu)}$ in the feedforward neural network, for $\ell = 1,2,\ldots, L$,  
the $j$th unit in the $\ell$th layer receives an affine signal from the $(\ell-1)$th layer as  
\begin{align}
u_{j,\mu}^{(\ell)} = \sum_{i=1}^{H_{\ell-1}}W_{j,i}^{(\ell)}z_{i,\mu}^{(\ell-1)},
\label{eqn:Affine_NN}
\end{align}
where $W_{j,i}^{(\ell)}$ is the directed-connection parameter from the $i$th unit in the $(\ell-1)$th layer to the $j$th unit in the $\ell$th layer, 
and $z_{i,\mu}^{(\ell-1)}$ is the output signal of the $i$th unit in the $(\ell-1)$th layer, 
where $z_{i,\mu}^{(0)}$ is identified as $\mrm{x}_{i, \mu}$. 
After receiving the affine signal, the $j$th unit in the $\ell$th layer outputs  
\begin{align}
z_{j,\mu}^{(\ell)} := a_{\ell}\big(b_j^{(\ell)} + u_{j,\mu}^{(\ell)}\big) 
\label{eqn:Output_NN}
\end{align}
to the upper layer, where $b_j^{(\ell)}$ is the bias parameter of the unit and $a_{\ell}(x)$ is the specific activation function of the $\ell$th layer. 
In the classification, for input $\mbf{x}^{(\mu)}$, the network output (the 1-of-$K$ vector) is usually determined through the $K$ class probabilities, $p_{1,\mu}, p_{2,\mu}, \ldots, p_{K,\mu}$, 
which are obtained through a softmax operation for $\{z_{k, \mu}^{(L)}\}$ ($H_L = K$): 
\begin{align}
p_{k,\mu} := \frac{\exp z_{k, \mu}^{(L)}}{\sum_{r=1}^K \exp z_{r, \mu}^{(L)}}.
\label{eqn:ClassProbability}
\end{align}
The network output is the 1-of-$K$ vector whose $k^*$th element is one, where $k^* = \argmax_k p_{k,\mu}$. 
The feedforward propagation is illustrated in Figure \ref{fig:DNN}.

Learning with BN is based on mini-batch wise stochastic gradient descent~\cite{BN2015}. 
In BN, the training dataset, $\mcal{D}$, is divided into $R$ mini-batches: $B_1,B_2,\ldots,B_R \subseteq \Omega$. 
In the following sections, the feedforward and back propagations during training in the $\ell$th layer for mini-batch $B_r$ are explained. 
Although all the signals appearing in the explanation depend on the index of mini-batch $r$,
the explicit description of the dependence of $r$ is omitted unless there is a particular reason.

\subsection{Feedforward propagation}

In the feedforward propagation for input $\mbf{x}^{(\mu)}$ $(\mu \in B_r)$ of the $\ell$th layer with BN,  
the form of $u_{j,\mu}^{(\ell)}$ in equation (\ref{eqn:Affine_NN}) is replaced by
\begin{align}
u_{j,\mu}^{(\ell)} = \gamma_j^{(\ell)}y_{j,\mu}^{(\ell)},\quad
y_{j,\mu}^{(\ell)} :=\frac{\lambda_{j,\mu}^{(\ell)} - m_{j}^{(\ell)}}{\big(v_{j}^{(\ell)} + \varepsilon \big)^{1/2}},
\quad \lambda_{j,\mu}^{(\ell)} := \sum_{i=1}^{H_{\ell-1}}W_{j,i}^{(\ell)}z_{i,\mu}^{(\ell-1)},
\label{eqn:Affine_NN_BN}
\end{align}
where 
\begin{align}
m_{j}^{(\ell)}=\frac{1}{|B_r|}\sum_{\mu \in B_r}\lambda_{j,\mu}^{(\ell)}
\label{eqn:affine_mean}
\end{align}
and
\begin{align}
v_{j}^{(\ell)}=\frac{1}{|B_r|-1}\sum_{\mu \in B_r}\big(\lambda_{j,\mu}^{(\ell)} - m_{j}^{(\ell)}\big)^2
\label{eqn:affine_variance}
\end{align}
are the mean and (unbiased) variance of the affine signals over $B_r$, respectively. Here, $|B_r|$ denotes the size of $B_r$; moreover, $|B_r| \geq 2$ is assumed.
The constant $\varepsilon$ in equation (\ref{eqn:Affine_NN_BN}) is set to a negligibly small positive value to avoid zero division; 
further, it is usually set to an infinitesimal values, e.g., $\varepsilon = 10^{-8}$. 
In BN, the distribution of the affine signals $\{\lambda_{j,\mu}^{(\ell)}\}$ over $B_r$ is standardized. 
This implies that the mean and (unbiased) variance of $\{y_{j,\mu}^{(\ell)}\}$ over $B_r$ are always zero and one (when $\varepsilon \to 0$), respectively: 
\begin{align}
\frac{1}{|B_r|}\sum_{\mu \in B_r}y_{j,\mu}^{(\ell)} = 0,\quad \frac{1}{|B_r|-1}\sum_{\mu \in B_r}(y_{j,\mu}^{(\ell)})^2 =1. 
\label{eqn:standardized_BN}
\end{align} 
The factors $\{\gamma_j^{(\ell)}\}$ in equation (\ref{eqn:Affine_NN_BN}) are also learning parameters with $\{b_j^{(\ell)}\}$ and $\{W_{j,i}^{(\ell)}\}$, 
which are determined using an appropriate back-propagation algorithm.

It is noteworthy that in the inference stage (after training), 
$M_{j}^{(\ell)}:=R^{-1}\sum_{r=1}^R m_{j,r}^{(\ell)}$ and $V_{j}^{(\ell)}:=(R-1)^{-1}\sum_{r=1}^R v_{j,r}^{(\ell)}$ (i.e., the moving average values over all the mini-batches) 
are usually used instead of $m_{j}^{(\ell)}$ and $v_{j}^{(\ell)}$, respectively, in equation (\ref{eqn:Affine_NN_BN}),  
where $m_{j,r}^{(\ell)}$ and $v_{j,r}^{(\ell)}$ indicate the mean and variance, respectively, computed for mini-batch $B_r$, 
namely, they are identified as equations (\ref{eqn:affine_mean}) and (\ref{eqn:affine_variance}), respectively. 

\subsection{Back propagation and gradients of loss}

In this section, the back-propagation rules with or without BN are explained. 
For expressing the back-propagation rules, a back-propagating signal of the $j$th unit in the $\ell$th layer for the $\mu$th data point ($\mu \in B_r$) is defined as
\begin{align}
\delta_{j,\mu}^{(\ell)}:=\frac{\partial E_{r}}{\partial u_{j,\mu}^{(\ell)}} \quad (\ell = 1,2,\ldots, L),
\label{eqn:def_DeltaSignal}
\end{align}
where $E_r \propto \sum_{\mu \in B_r}e_{\mu}$ is the loss over mini-batch $B_r$, and 
$e_{\mu}$ is the loss for the $\mu$th data point, i.e., $e_{\mu} = f(\mbf{x}_{\mu}, \mbf{t}_{\mu}; \theta)$ for equation (\ref{eqn:LossFunction}) 
or $e_{\mu} = w_{\mu}f(\mbf{x}_{\mu}, \mbf{t}_{\mu}; \theta)$ for equation (\ref{eqn:WeightedLossFunction}).

First, consider the back-propagation rule for the $\ell$th layer ``without'' BN, i.e., the case in which the form of $u_{j,\mu}^{(\ell)}$ is defined by equation (\ref{eqn:Affine_NN}). 
In this case, the back-propagation rule is obtained by using the chain rule and equations (\ref{eqn:Affine_NN}) and (\ref{eqn:Output_NN}), i.e., 
\begin{align}
\delta_{j,\mu}^{(\ell)} =\sum_{k = 1}^{H_{\ell + 1}}\sum_{\alpha \in B_r} \frac{\partial E_{r}}{\partial u_{k,\alpha}^{(\ell + 1)}}\frac{\partial u_{k,\alpha}^{(\ell + 1)}}{\partial u_{j,\mu}^{(\ell)}}
=\sum_{k = 1}^{H_{\ell + 1}} \delta_{k,\mu}^{(\ell + 1)}\frac{\partial u_{k,\mu}^{(\ell + 1)}}{\partial u_{j,\mu}^{(\ell)}}
= a_{\ell}^{(1)}(u_{j,\mu}^{(\ell)})\sum_{k = 1}^{H_{\ell + 1}}W_{k,j}^{(\ell + 1)}\delta_{k,\mu}^{(\ell + 1)},
\label{eqn:BackPropagation_standard}
\end{align}
for $\ell = 1,2,\ldots, L-1$, where $a_{\ell}^{(1)}(x) := d a_{\ell}(x) / dx$. 
Note that $\delta_{j,\mu}^{(L)}$ is obtained by directly differentiating $E_r$ by $u_{j,\mu}^{(L)}$. 
By using the back-propagating signals, the gradients of $E_r$ with respect to $b_j^{(\ell)}$ and $W_{j,i}^{(\ell)}$ 
are expressed as
\begin{align}
\frac{\partial E_{r}}{\partial b_{j}^{(\ell)}} &= \sum_{k = 1}^{H_{\ell}}\sum_{\mu \in B_r}
 \frac{\partial E_{r}}{\partial u_{k,\mu}^{(\ell)}}\frac{\partial u_{k,\mu}^{(\ell)}}{\partial b_{j}^{(\ell)}} 
= \sum_{\mu \in B_r}\delta_{j,\mu}^{(\ell)}, 
\label{eqn:grad_b_standard}\\
\frac{\partial E_{r}}{\partial W_{j,i}^{(\ell)}} &= \sum_{k = 1}^{H_{\ell}}\sum_{\mu \in B_r}
 \frac{\partial E_{r}}{\partial u_{k,\mu}^{(\ell)}}\frac{\partial u_{k,\mu}^{(\ell)}}{\partial W_{j,i}^{(\ell)}} 
= \sum_{\mu \in B_r}\delta_{j,\mu}^{(\ell)} z_{i,\mu}^{(\ell-1)},
\label{eqn:grad_W_standard}
\end{align}
respectively, for $\ell = 1,2,\ldots, L$.

Next, consider the expression of the back-propagation rule for the $\ell$th layer ``with'' BN, 
i.e., the case in which the form of $u_{j,\mu}^{(\ell)}$ is defined by equation (\ref{eqn:Affine_NN_BN}). 
Using a similar manipulation as equation (\ref{eqn:BackPropagation_standard}) (i.e., using the chain rule and equations (\ref{eqn:Output_NN}) and (\ref{eqn:Affine_NN_BN})), 
\begin{align}
\delta_{j,\mu}^{(\ell)} 
&=\sum_{k = 1}^{H_{\ell + 1}}\sum_{\alpha \in B_r} \frac{\partial E_{r}}{\partial u_{k,\alpha}^{(\ell + 1)}}\frac{\partial u_{k,\alpha}^{(\ell + 1)}}{\partial u_{j,\mu}^{(\ell)}}
=\sum_{k = 1}^{H_{\ell + 1}} \sum_{\alpha \in B_r} \delta_{k,\alpha}^{(\ell + 1)}\frac{\partial u_{k,\alpha}^{(\ell + 1)}}{\partial u_{j,\mu}^{(\ell)}}\nn
&= a_{\ell}^{(1)}(u_{j,\mu}^{(\ell)})\sum_{k = 1}^{H_{\ell + 1}}W_{k,j}^{(\ell + 1)}
\bigg\{  D_{k,\mu}^{(\ell + 1)} - \frac{1}{|B_r|}\sum_{\alpha \in B_r}D_{k,\alpha}^{(\ell + 1)}
\Big(1 + \frac{|B_r|}{|B_r| - 1}y_{k,\alpha}^{(\ell + 1)}y_{k,\mu}^{(\ell + 1)}\Big)\bigg\}
\label{eqn:BackPropagation_BN}
\end{align}
is obtained for $\ell = 1,2,\ldots, L-1$, 
where
\begin{align*}
D_{j, \mu}^{(\ell)}:=\frac{\gamma_j^{(\ell)} }{\big(v_{j}^{(\ell)} + \varepsilon \big)^{1/2}}\delta_{j,\mu}^{(\ell)}.
\end{align*}
In this case, the gradients of $E_r$ with respect to $b_j^{(\ell)}$, $\gamma_j^{(\ell)}$, and $W_{j,i}^{(\ell)}$ are as follows.  
The gradient $\partial E_r / \partial b_{j}^{(\ell)}$ has the same expression as equation (\ref{eqn:grad_b_standard}).
The gradients with respect to $\gamma_j^{(\ell)}$ and $W_{j,i}^{(\ell)}$ are
\begin{align}
\frac{\partial E_{r}}{\partial \gamma_{j}^{(\ell)}} &= \sum_{k = 1}^{H_{\ell}}\sum_{\mu \in B_r}
 \frac{\partial E_{r}}{\partial u_{k,\mu}^{(\ell)}}\frac{\partial u_{k,\mu}^{(\ell)}}{\partial \gamma_{j}^{(\ell)}} 
= \sum_{\mu \in B_r}\delta_{j,\mu}^{(\ell)}y_{k,\mu}^{(\ell)}, 
\label{eqn:grad_gamma_BN}\\
\frac{\partial E_{r}}{\partial W_{j,i}^{(\ell)}} &= \sum_{k = 1}^{H_{\ell}}\sum_{\mu \in B_r}
 \frac{\partial E_{r}}{\partial u_{k,\mu}^{(\ell)}}\frac{\partial u_{k,\mu}^{(\ell)}}{\partial W_{j,i}^{(\ell)}} 
= \sum_{\mu \in B_r}\bigg\{  D_{j,\mu}^{(\ell)} - \frac{1}{|B_r|}\sum_{\alpha \in B_r}D_{j,\alpha}^{(\ell)}
\Big(1 + \frac{|B_r|}{|B_r| - 1}y_{j,\alpha}^{(\ell)}y_{j,\mu}^{(\ell)}\Big)\bigg\}z_{i,\mu}^{(\ell-1)},
\label{eqn:grad_W_BN}
\end{align}
respectively, for $\ell = 1,2,\ldots, L$.

\subsection{Numerical experiment}
\label{sec:experiment1}

For the experiments presented in this section, three different datasets were used: MNIST, Fashion-MNIST~\cite{fashion-MNIST2017}, and CIFAR-10. 
MNIST is a dataset of handwritten digit images, i.e., $0, 1,\ldots,9$, which consists of a training set of 60000 data points and a test set of 10000 data points. 
Each data point in MNIST consists of the input image, i.e., a $28 \times 28$ grayscale digit image, and the corresponding target digit label. 
Fashion-MNIST is a dataset of Zalando's article images consisting of a training set of 60000 data points and a test set of 10000 data points,  
in which each data point consists of the input image, i.e., a $28 \times 28$ grayscale image, 
and the corresponding target article label from 10 classes, such as ``t-shirt,'' ``trouser,'' ``pullover.'' 
CIFAR-10 is a dataset of color images of 10 different objects, consisting of a training set of 50000 data points and test set of 10000 data points, 
in which each data point consists of the input image, i.e., a $32 \times 32$ RGB color image, 
and the corresponding target article label from 10 classes, such as ``airplane,'' ``automobile,'' ``bird.''.  
The color input images in CIFAR-10 were converted to grayscale images in the following experiments.

For the three different datasets, some \textit{imbalanced} classification problems were considered. 
For MNIST, a four-class classification problem of ``three,'' ``four,'' ``seven,'',  and ``nine'' was considered.  
For Fashion-MNIST, a three-class classification problem of ``sandal,'' ``sneaker,'', and ``ankle boot'' was considered.
For CIFAR-10, a three-class classification problem of ``airplane,'' ``deer,'' and ``truck'' was considered. 
The number of training and test data points used in these experiments are listed in Tables \ref{tab:number_of_data_MNIST}--\ref{tab:number_of_data_CIFAR10}. 
The data points in the imbalanced training sets were randomly picked up from the original training sets 
and those in the test sets were all data points of the corresponding labels in the original test sets.  
All the inputs were normalized by dividing by 255 during the preprocessing. 

For the experiments, a four-layered feedforward neural network ($L = 3$) was used in which the sizes of the first and second hidden layers were 300.
The activation functions of the first and second hidden layers were $\tanh(x)$ and that of the last hidden layer was the identity function. 
The network output, $\bm{t} \in T_K$, was computed from the output signals of the last hidden layer using the softmax operation, as explained in the first part of this section.
BN was adopted for the first and second hidden layers.
In the training, the Xavier initialization~\cite{Xavier2010}, Adamax optimizer~\cite{Adam2015} with a mini-batch size of 128, 
and cross-entropy loss, $f(\mbf{x}_{\mu}, \mbf{t}_{\mu}; \theta) = -\sum_{k=1}^K \mrm{t}_{k,\mu} \ln p_{k,\mu}$, 
were used, where $p_{k,\mu}$ is the class probability defined in equation (\ref{eqn:ClassProbability}). 
The setting of the hyperparameters in the Adamax optimizer followed that recommended in Reference~\cite{Adam2015}. 
Each mini-batch loss, $E_r \propto \sum_{\mu \in B_r}e_{\mu}$, was normalized by dividing by the effective size of the corresponding mini-batch. 
The effective size of $B_r$ was $|B_r|$ for the non weighted loss function in equation (\ref{eqn:LossFunction}) and was $Z_r$ for WLF, where
\begin{align}
Z_r :=\sum_{\mu \in B_r}w_{\mu}.
\label{eqn:definition_normalization-factor_Zr}
\end{align}

\begin{table}[t]
\centering
\caption{Number of data points used in the experiment using MNIST.} 
\label{tab:number_of_data_MNIST}
\begin{tabular}{c|c|c|c|c|}
\cline{2-5}
 & 3 (minority) & 4 (minority) & 7 (minority) & 9 (majority)\\ \hline
\multicolumn{1}{|c|}{train} & 5 & 5 & 5 & 5000 \\ \hline
\multicolumn{1}{|c|}{test} & 1010 & 982 & 1028 & 1009 \\ \hline
\end{tabular}
\end{table}

\begin{table}[t]
\centering
\caption{Number of data points used in the experiment using Fashion-MNIST. }
\label{tab:number_of_data_fashion-MNIST}
\begin{tabular}{c|c|c|c|}
\cline{2-4}
 & sandal (minority) & sneaker (minority) & ankle boot (majority)  \\ \hline
\multicolumn{1}{|c|}{train} & 5 & 5 & 5000  \\ \hline
\multicolumn{1}{|c|}{test} & 1000 & 1000 & 1000  \\ \hline
\end{tabular}
\end{table}

\begin{table}[t]
\centering
\caption{Number of data points used in the experiment using CIFAR-10. }
\label{tab:number_of_data_CIFAR10}
\begin{tabular}{c|c|c|c|}
\cline{2-4}
 & airplane (minority) & deer (minority) & truck (majority)  \\ \hline
\multicolumn{1}{|c|}{train} & 20 & 30 & 4500  \\ \hline
\multicolumn{1}{|c|}{test} & 1000 & 1000 & 1000  \\ \hline
\end{tabular}
\end{table}

The results of the experiments are shown in Tables \ref{tab:classification_rate_MNIST_ex1}--\ref{tab:classification_rate_CIFAR10_ex1}. 
For each dataset, three different methods were used: 
(a) the standard loss function, i.e., $\mcal{L}(\theta)$, combined with BN (LF+BN), 
(b) WLF, i.e., $\mcal{L}_{\mrm{w}}(\theta)$, with ICF combined with BN (WLF(ICF)+BN), 
and (c) WLF with CBL ($\beta = (N-1) / N$) combined with BN (WLF(CBL)+BN).   
The accuracies listed in the tables are the average values over 30 experiments 
(in all the experiments, the training datasets were randomly reselected for each experiment). 
In each experiment, the models obtained after 200 epochs of training were used for the tests.
The classification accuracies for the majority classes were significantly good and those for the minority classes were very poor as expected. 
However, the results of (b) and (c) were not sufficiently improved when compared to those of (a). 
This indicates that the correction for the effective data size in WLF does not sufficiently function in the present experiments. 

In CBL, although the value of hyperparameter $\beta$ can be tuned, the fixed value, $\beta = (N-1) / N$, was used in the present experiments. 
The results of (c) could be improved by optimizing the value of $\beta$.
However, this is not essential for the primary claim of this study.
Because the aim of this study is to resolve the size-inconsistency problem and improve the classification performance, 
the difference in the baseline is not important.

\begin{table}[t]
\centering
\caption{Classification accuracies (of each digit and overall) for the test set in MNIST.}
\label{tab:classification_rate_MNIST_ex1}
\begin{tabular}{c|c|c|c|c||c|}
\cline{2-6}
 & 3 (minority) & 4 (minority) & 7 (minority) & 9 (majority)  & overall\\ \hline
\multicolumn{1}{|l|}{(a) LF+BN} & 13.7\% & 3.4\% & 6.0\% & 100.0\% & 30.8\% \\ \hline
\multicolumn{1}{|l|}{(b) WLF(ICF)+BN} & 15.3\% & 4.8\% & 9.9\% & 100.0\% & 32.6\% \\ \hline
\multicolumn{1}{|l|}{(c) WLF(CBL)+BN} & 14.7\% & 4.6\% & 9.9\% & 100.0\% & 32.4\% \\ \hline
\end{tabular}
\end{table}

\begin{table}[t]
\centering
\caption{Classification accuracies (of each article and overall) for the test set in Fashion-MNIST.}
\label{tab:classification_rate_fashion-MNIST_ex1}
\begin{tabular}{c|c|c|c||c|}
\cline{2-5}
 & sandal (minority) & sneaker (minority) & ankle boot (majority)  & overall\\ \hline
\multicolumn{1}{|l|}{(a) LF+BN} & 6.6\% & 21.4\% & 100.0\% &  42.7\% \\ \hline
\multicolumn{1}{|l|}{(b) WLF(ICF)+BN} & 8.3\% & 25.2\% & 100.0\% & 44.5\% \\ \hline
\multicolumn{1}{|l|}{(c) WLF(CBL)+BN} & 8.1\% & 25.7\% & 100.0\% & 44.6\% \\ \hline
\end{tabular}
\end{table}

\begin{table}[t]
\centering
\caption{Classification accuracies (of each object and overall) for the test set in CIFAR-10.}
\label{tab:classification_rate_CIFAR10_ex1}
\begin{tabular}{c|c|c|c||c|}
\cline{2-5}
 & airplane (minority) & deer (minority) & truck (majority)  & overall\\ \hline
\multicolumn{1}{|l|}{(a) LF+BN} & 2.3\% & 2.1\% & 99.9\% &  34.8\% \\ \hline
\multicolumn{1}{|l|}{(b) WLF(ICF)+BN} & 4.6\% & 4.3\% & 99.8\% & 36.2\% \\ \hline
\multicolumn{1}{|l|}{(c) WLF(CBL)+BN} & 4.6\% & 4.1\% & 99.8\% & 36.2\% \\ \hline
\end{tabular}
\end{table}

\section{Weighted Batch Normalization (WBN)}
\label{sec:WBN}

As presented in Section \ref{sec:experiment1}, the simple combination of WLF and BN appears to be an inferior method. 
This is considered to be due to the inconsistency of the effective data size in WLF and BN. 
As mentioned in Section \ref{sec:WLF}, in WLF, 
the interpretation of the effective sizes of data points is changed in accordance with the corresponding weights. 
However, in BN, one data point is treated as just one data point in the computation of the mean and variance of the affine signals (cf. equations (\ref{eqn:affine_mean}) and (\ref{eqn:affine_variance})).
This causes a size-inconsistency problem, which inhibits the effect of the correction for the effective data size in WLF.
To resolve this problem, a modified BN (referred to as \textit{weighted batch normalization} (WBN)) is proposed for WLF. 

\subsection{Feedforward and back propagations for WBN}

The idea of the proposed method is simple and natural. 
To maintain consistency with the interpretation of data size, the sizes of the mini-batches should be reviewed according to the corresponding weights in BN. 
This implies that 
\begin{align}
m_{j}^{(\ell)}=\frac{1}{Z_{r}}\sum_{\mu \in B_r} w_{\mu} \lambda_{j,\mu}^{(\ell)}
\label{eqn:affine_expectation_WBN}
\end{align}
and
\begin{align}
v_{j}^{(\ell)}=\frac{1}{G_r}\sum_{\mu \in B_r}w_{\mu}\big(\lambda_{j,\mu}^{(\ell)} - m_{j}^{(\ell)}\big)^2
\label{eqn:affine_variance_WBN}
\end{align}
should be used in equation (\ref{eqn:Affine_NN_BN}) instead of equations (\ref{eqn:affine_mean}) and (\ref{eqn:affine_variance}), 
where
\begin{align}
G_r := \frac{1}{Z_r}\Big( Z_r^2 - \sum_{\mu \in B_r}w_{\mu}^2\Big)
\label{eqn:definition_normalization-factor_Gr}
\end{align}
is the normalization factor for the variance. 
The normalization factor for the mean, $Z_r$, is already defined in Equation (\ref{eqn:definition_normalization-factor_Zr}).
Equations (\ref{eqn:affine_expectation_WBN}) and (\ref{eqn:affine_variance_WBN}) are the modified versions of  equations (\ref{eqn:affine_mean}) and (\ref{eqn:affine_variance}), 
weighted in the same manner as WLF. 
The normalization factors in equations (\ref{eqn:affine_expectation_WBN}) and (\ref{eqn:affine_variance_WBN}) ensure that they are unbiased; refer Appendix \ref{app:Unbiassness}. 
In the previous study~\cite{NOLTA2019}, the normalization factor $G_r$ was defined by $Z_r - 1$. 
However, such a definition is inappropriate from the perspective of unbiasedness of $v_{j}^{(\ell)}$.
When $w_{\mu}$ is a constant, WBN is equivalent to BN, 
because equations (\ref{eqn:affine_expectation_WBN}) and (\ref{eqn:affine_variance_WBN}) are reduced to equations (\ref{eqn:affine_mean}) and (\ref{eqn:affine_variance}) in this case. 

In BN, the distribution of $\{y_{j,\mu}^{(\ell)}\}$ over $B_r$ is standardized, as shown in equation (\ref{eqn:standardized_BN}). 
Conversely, in WBN, it is not standardized; however, its weighted distribution is standardized: 
\begin{align*}
\frac{1}{Z_r}\sum_{\mu \in B_r} w_{\mu} y_{j,\mu}^{(\ell)} = 0,\quad
\frac{1}{G_r}\sum_{\mu \in B_r} w_{\mu} (y_{j,\mu}^{(\ell)})^2 = 1,
\end{align*}
when $\varepsilon \to 0$. 
This standardization property reflects the effective data size in WLF. 
In WLF, the $\mu$th data point is replicated according to the corresponding weight as mentioned in Section \ref{sec:WLF}; 
therefore, the corresponding signal $y_{j,\mu}^{(\ell)}$ should also be replicated in the same manner. 
The above standardization property implies this notion.

In WBN, the back-propagation rule in equation (\ref{eqn:BackPropagation_BN}) is modified as
\begin{align}
\delta_{j,\mu}^{(\ell)} 
= a_{\ell}^{(1)}(u_{j,\mu}^{(\ell)})\sum_{k = 1}^{H_{\ell + 1}}W_{k,j}^{(\ell + 1)}
\bigg\{  D_{k,\mu}^{(\ell + 1)} - \frac{ w_{\mu}}{Z_r}\sum_{\alpha \in B_r}D_{k,\alpha}^{(\ell + 1)}
\Big(1 + \frac{Z_r}{G_r}y_{k,\alpha}^{(\ell + 1)}y_{k,\mu}^{(\ell + 1)}\Big)\bigg\}.
\label{eqn:BackPropagation_WBN}
\end{align}
The expressions for the gradients of $E_r$ with respect to $b_j^{(\ell)}$ and $\gamma_j^{(\ell)}$ are the same as equations (\ref{eqn:grad_b_standard}) and (\ref{eqn:grad_gamma_BN}), respectively. 
The gradient with respect to $W_{j,i}^{(\ell)}$ in equation (\ref{eqn:grad_W_BN}) is modified as
\begin{align}
\frac{\partial E_{r}}{\partial W_{j,i}^{(\ell)}} 
= \sum_{\mu \in B_r}\bigg\{  D_{j,\mu}^{(\ell)} - \frac{w_{\mu}}{Z_r}\sum_{\alpha \in B_r}D_{j,\alpha}^{(\ell)}
\Big(1 + \frac{Z_r}{G_r}y_{j,\alpha}^{(\ell)}y_{j,\mu}^{(\ell)}\Big)\bigg\}z_{i,\mu}^{(\ell-1)}.
\label{eqn:grad_W_WBN}
\end{align}

\subsection{Numerical experiment}
\label{sec:experiment2}

In this section, the experimental results of the proposed method for the same imbalanced classification problems as Section \ref{sec:experiment1} are presented. 
The detailed setting of the experiments was identical to that of the experiments presented in Section \ref{sec:experiment1}. 
For each experiment, two different methods were used: 
(d) WLF with ICF combined with WBN (WLF(ICF)+WBN) and (e) WLF with CBL ($\beta = (N-1) / N$) combined with WBN (WLF(CBL)+WBN).  
Tables \ref{tab:classification_rate_MNIST_ex2}--\ref{tab:classification_rate_CIFAR10_ex2} list the results of the experiments 
(for comparison, the results of (a)--(c) in Tables \ref{tab:classification_rate_MNIST_ex1}--\ref{tab:classification_rate_CIFAR10_ex1} are presented again).
It is observed that the classification accuracies for the minority classes in all the experiments were noticeably improved. 
This implies that the size-inconsistency problem is resolved by WBN and consequently the correction for the effective data size in WLF functions well, as expected.

The same experiments using WBN, as proposed in the previous study were executed, in which $G_r$ was defined by $Z_r - 1$~\cite{NOLTA2019}. 
The results obtained from these experiments were marginally worse than those listed in Tables \ref{tab:classification_rate_MNIST_ex2}--\ref{tab:classification_rate_CIFAR10_ex2}. 
This implies that the definition of $G_r$ in equation (\ref{eqn:definition_normalization-factor_Gr}) is better from the perspective of  
not only the unbiasedness of the variance $v_j^{(\ell)}$ but also of the performance of classification.

\begin{table}[t]
\centering
\caption{Classification accuracies (of each digit and overall) for the test set in MNIST.}
\label{tab:classification_rate_MNIST_ex2}
\begin{tabular}{c|c|c|c|c||c|}
\cline{2-6}
 & 3 (minority) & 4 (minority) & 7 (minority) & 9 (majority)  & overall\\ \hline
\multicolumn{1}{|l|}{(a) LF+BN} & 13.7\% & 3.4\% & 6.0\% & 100.0\% & 30.8\% \\ \hline
\multicolumn{1}{|l|}{(b) WLF(ICF)+BN} & 15.3\% & 4.8\% & 9.9\% & 100.0\% & 32.6\% \\ \hline
\multicolumn{1}{|l|}{(c) WLF(CBL)+BN} & 14.7\% & 4.6\% & 9.9\% & 100.0\% & 32.4\% \\ \hline\hline
\multicolumn{1}{|l|}{(d) WLF(ICF)+WBN} & 57.4\% & 40.3\% & 42.9\% & 99.5\% & 60.1\% \\ \hline
\multicolumn{1}{|l|}{(e) WLF(CBL)+WBN} & 52.2\% & 36.9\% & 37.7\% & 99.7\% & 56.7\% \\ \hline
\end{tabular}
\end{table}

\begin{table}[t]
\centering
\caption{Classification accuracies (of each article and overall) for the test set in Fashion-MNIST.}
\label{tab:classification_rate_fashion-MNIST_ex2}
\begin{tabular}{c|c|c|c||c|}
\cline{2-5}
 & sandal (minority) & sneaker (minority) & ankle boot (majority)  & overall\\ \hline
\multicolumn{1}{|l|}{(a) LF+BN} & 6.6\% & 21.4\% & 100.0\% &  42.7\% \\ \hline
\multicolumn{1}{|l|}{(b) WLF(ICF)+BN} & 8.3\% & 25.2\% & 100.0\% & 44.5\% \\ \hline
\multicolumn{1}{|l|}{(c) WLF(CBL)+BN} & 8.1\% & 25.7\% & 100.0\% & 44.6\% \\ \hline \hline
\multicolumn{1}{|l|}{(d) WLF(ICF)+WBN} & 25.7\% & 40.4\% & 99.1\% & 55.1\% \\ \hline
\multicolumn{1}{|l|}{(e) WLF(CBL)+WBN} & 23.3\% & 38.9\% & 99.8\% & 54.0\% \\ \hline
\end{tabular}
\end{table}

\begin{table}[t]
\centering
\caption{Classification accuracies (of each object and overall) for the test set in CIFAR-10.}
\label{tab:classification_rate_CIFAR10_ex2}
\begin{tabular}{c|c|c|c||c|}
\cline{2-5}
 & airplane (minority) & deer (minority) & truck (majority)  & overall\\ \hline
\multicolumn{1}{|l|}{(a) LF+BN} & 2.3\% & 2.1\% & 99.9\% &  34.8\% \\ \hline
\multicolumn{1}{|l|}{(b) WLF(ICF)+BN} & 4.6\% & 4.3\% & 99.8\% & 36.2\% \\ \hline
\multicolumn{1}{|l|}{(c) WLF(CBL)+BN} & 4.6\% & 4.1\% & 99.8\% & 36.2\% \\ \hline \hline
\multicolumn{1}{|l|}{(d) WLF(ICF)+WBN} & 18.6\% & 13.0\% & 95.0\% & 42.1\% \\ \hline
\multicolumn{1}{|l|}{(e) WLF(CBL)+WBN} & 11.7\% & 11.3\% & 96.3\% & 39.8\% \\ \hline
\end{tabular}
\end{table}

\section{Summary and Future Works}
\label{sec:summary}

The present paper proposed a new BN method for learning based on WLF. 
The idea of the proposed method is simple but essential. 
The proposed BN, i.e., WBN, can resolve the size-inconsistency problem which arises in the combination of WLF and BN; moreover,  
it improved the classification performance in data-imbalanced environments, as demonstrated in the numerical experiments. 
WBN on standard feedforward neural networks was validated for three different datasets (MNIST, Fashion-MNIST, and CIFAR-10)  
and for two different weight settings (ICF and CBL) through the numerical experiments. 
However, the present paper focused on only standard feedforward neural networks; therefore,  
further experiments are required to demonstrate the efficacy of WBN for more practical applications; 
for example, WBN for a larger network based on a convolution neural network such as a wide residual network~\cite{WResNet2016} 
and WBN for other more difficult datasets such as ImageNet should be investigated.  
Furthermore, deepening the mathematical aspect of WBN, such as the internal covariance shift in WBN, is also important.
These are important aspects that will be studied in our future project.

Recently, certain extensions of BN were proposed such as transferable normalization~\cite{TransBN2019} and domain-specific batch normalization~\cite{DomSpeBN2019}.  
The proposed method can be applied to these extension methods by modifying the computations of the means and variances of the mini-batches  
in these methods in a manner similar to that considered in equations (\ref{eqn:affine_expectation_WBN}) and (\ref{eqn:affine_variance_WBN}).
Only the classification problem is considered in this study. 
However, the size-inconsistency problem in the combination of WLF and BN will also arise in other types of problems, e.g., regression problem. 
WBN must be applicable to these cases because the idea of WBN is independent of the style of output. 
The application of the proposed method to other types of BNs and problems will also be addressed in our future studies.

\section*{\normalsize{acknowledgments}}

This work was partially supported by JSPS KAKENHI (Grant Numbers 15H03699, 18K11459, and 18H03303), 
JST CREST (Grant Number JPMJCR1402), and the COI Program from the JST (Grant Number JPMJCE1312).  

\appendix

\section{Unbiasedness of Estimators in Equations (\ref{eqn:affine_expectation_WBN}) and (\ref{eqn:affine_variance_WBN})}
\label{app:Unbiassness}

In this appendix, it is shown that the mean and variance in equations (\ref{eqn:affine_expectation_WBN}) and (\ref{eqn:affine_variance_WBN}), are unbiased. 
By omitting indices ($j$ and $\ell$) unrelated to this analysis, they are expressed as
\begin{align}
m(\bm{\lambda})=\frac{1}{Z_{r}}\sum_{\mu \in B_r} w_{\mu} \lambda_{\mu}, \quad 
v(\bm{\lambda})=\frac{1}{G_r}\sum_{\mu \in B_r}w_{\mu}\big(\lambda_{\mu} - m(\bm{\lambda})\big)^2
\label{eqn:def_m&v_app}
\end{align}
Here, assume that $\{\lambda_{\mu}\}$ are i.i.d. samples drawn from a distribution $p(\lambda)$. 
The mean and variance of $p(\lambda)$ are denoted by $\mu_{\mrm{true}}$ and $\sigma_{\mrm{true}}^2$, respectively.
The expectation of $m(\bm{\lambda})$ in equation (\ref{eqn:def_m&v_app}) over $\prod_{\mu \in B_r} p(\lambda_{\mu})$ is 
\begin{align*}
\int m(\bm{\lambda}) \prod_{\mu \in B_r} p(\lambda_{\mu}) d \lambda_{\mu} = \frac{1}{Z_{r}}\sum_{\mu \in B_r} w_{\mu} \int \lambda_{\mu}p(\lambda_{\mu}) d \lambda_{\mu} 
= \mu_{\mrm{true}}.
\end{align*}
Therefore, $m(\bm{\lambda})$ is the unbiased estimator. 
Similarly, the expectation of $v(\bm{\lambda})$ in equation (\ref{eqn:def_m&v_app}) is 
\begin{align*}
&\int v(\bm{\lambda}) \prod_{\mu \in B_r} p(\lambda_{\mu}) d \lambda_{\mu} 
= \frac{1}{G_{r}}\sum_{\mu \in B_r} w_{\mu} \int\big(\lambda_{\mu} - m\big)^2 \prod_{\alpha \in B_r}p(\lambda_{\alpha}) d \lambda_{\alpha} \nn
&= \frac{1}{G_{r}}\sum_{\mu \in B_r} w_{\mu} \int \Big(\lambda_{\mu}^2 -\frac{2}{Z_r}\sum_{\nu \in B_r}w_{\nu}\lambda_{\nu} \lambda_{\mu}
+ \frac{1}{Z_r^2}\sum_{\nu, \tau \in B_r}w_{\nu}w_{\tau}\lambda_{\nu}\lambda_{\tau}\Big)\prod_{\alpha \in B_r}p(\lambda_{\alpha}) d \lambda_{\alpha}
=\sigma_{\mrm{true}}^2.
\end{align*}
Here,  
\begin{align*}
\sum_{\nu \in B_r} w_{\nu} \int\lambda_{\nu} \lambda_{\mu}\prod_{\alpha \in B_r}p(\lambda_{\alpha}) d \lambda_{\alpha} &= 
Z_r \mu_{\mrm{true}}^2 + w_{\mu} \sigma_{\mrm{true}}^2,\nn
\sum_{\nu, \tau \in B_r}w_{\nu}w_{\tau}\int\lambda_{\nu} \lambda_{\tau}\prod_{\alpha \in B_r}p(\lambda_{\alpha}) d \lambda_{\alpha}
&=Z_r^2 \mu_{\mrm{true}}^2 + \sum_{\nu \in B_r} w_{\nu}^2\sigma_{\mrm{true}}^2
\end{align*}
are used. Therefore, $v(\bm{\lambda})$ is also an unbiased estimator.

\bibliographystyle{unsrt}
\bibliography{citation}

\end{document}